\documentclass[10pt, a4paper]{article}
\usepackage{lrec2022} % this is the new LREC2022 Style
\usepackage{multibib}
\newcites{languageresource}{Language Resources}
\usepackage{graphicx}
\usepackage{tabularx}
\usepackage{soul}
\usepackage{covington}
\usepackage[utf8]{inputenc}
\usepackage[T1]{fontenc}
\PassOptionsToPackage{hyphens}{url}\usepackage{hyperref}

\usepackage{titlesec}
\titleformat{\section}{\normalfont\large\bfseries\center}{\thesection.}{1em}{}
\titleformat{\subsection}{\normalfont\SmallTitleFont\bfseries\raggedright}{\thesubsection.}{1em}{}
\titleformat{\subsubsection}{\normalfont\normalsize\bfseries\raggedright}{\thesubsubsection.}{1em}{}
\renewcommand\thesection{\arabic{section}}
\renewcommand\thesubsection{\thesection.\arabic{subsection}}
\renewcommand\thesubsubsection{\thesubsection.\arabic{subsubsection}}
\usepackage{epstopdf}
\usepackage[utf8]{inputenc}

\usepackage{hyperref}
\usepackage{xstring}

\usepackage{bm}

\usepackage{color}

\newcommand\ExLi[4]{
\item[#1$_{#2}$] #3
\vspace{-3mm}
\item[#1$_{#2}$] #4
}

\title{A Multi-Party Dialogue Ressource in French \\ \vspace*{.5\baselineskip}} 

\name{Maria Boritchev, Maxime Amblard} 

\address{Institute of Mathematics of the Polish Academy of Sciences
Warsaw, Poland\\ 
LORIA, UMR 7503, Université de Lorraine, CNRS, Inria, 54000 Nancy, France \\
         mboritchev@impan.pl, maxime.amblard@loria.fr\\}

\abstract{We present \textit{Dialogues in Games} (DinG), a corpus of manual transcriptions of real-life, oral, spontaneous multi-party dialogues between French-speaking players of the board game \textit{Catan}. Our objective is to make available a quality resource for French, composed of long dialogues, to facilitate their study in the style of~\cite{asher:hal-02124399}. In a general dialogue setting, participants share personal information, which makes it impossible to disseminate the resource freely and openly. In DinG, the attention of the participants is focused on the game, which prevents them from talking about themselves. In addition, we are conducting a study on the nature of the questions in dialogue, through annotation~\cite{blandon2019toward}, in order to develop more natural automatic dialogue systems.
 \\ \newline  }

\begin{document}

\maketitleabstract

\section{Introduction}

We envision our corpus as a corpus of spontaneous dialogues in French with a quality transcription. Its nature allows for large dissemination and high cross-domain reusability. Its length allows for a study from different perspectives. 
As the cost of producing resources is very high, we designed our corpus aiming for the widest possible use.
Furthermore, we want to present it as a corpus that follows the good practices of corpus collection, usable for the collection of other corpora, especially ones that cannot be disseminated largely because of the nature of the data they contain. 

One of the main inspirations in the design of DinG was the STAC project\footnote{\url{https://www.irit.fr/STAC/}} corpus~\cite{asher:hal-02124399}. The corpus of STAC is composed of chat logs from an online version of the board game \textit{Catan}, played by English speakers. STAC is annotated in Segmented Discourse Representation Theory (SDRT,~\cite{asher2003logics}). The proximity our two corpora share is a great opportunity to progress on comparisons between written and oral discourse on the same topics while opening on dialogue.

DinG is now used to study questions in dialogue in approaches such as \cite{boritchev:hal-02930715,boritchev2021dialogue}. As DinG is composed of long human-human interactions, it can be very largely used for dialogue studies in numerous fields. 

\paragraph{Links with other Studies of Dialogue}
Another motivation for developing DinG is to showcase the work we are conducting on highly sensible health-related data. Studying DinG can help not to tag a phenomenon as a conversational disorder by acknowledging the fact that this phenomenon occurs in a non-pathological setting as well. Data from DinG is also used in the Sémagramme team (LORIA, Inria Nancy Grand-Est) to help develop and showcase SLODiM\footnote{\url{https://slodim.loria.fr/}}, a tool for dialogue analysis in a medical (psychiatric) setting, developed in the line of Schizophrenia and Language, Analysis and Modeling\footnote{\url{https://team.inria.fr/semagramme/fr/slam/}} (SLAM)~\cite{amblard:hal-00955660},~ \cite{amblard:hal-01079308},~\cite{amblard:hal-01054391}.
The ODiM corpus follows the same transcription process as DinG. For the sake of science reproducibility, we developed DinG as a free-to-share corpus to showcase our studies, our tools, and our formal models. 

To work on dialogic interaction, we need to study the dynamics of dialogic exchanges. Thus, we wish to have the transcription, but also the time codes corresponding to the dialogue. We have therefore developed a fine segmentation in addition to the transcription.

\paragraph{Link with other Researches with Catan}

The corpus was designed to study human-human dialogue based on attested, spontaneous, and unconstrained oral data in French. We want to study different types of dialogue-encountered phenomena; on one hand, the mechanisms underlying the combination of dialogue turns, in order to produce a computational model of dialogue, on the other hand, the dynamics of interactions, in order to account for meta-levels of human-human interactions.

The proximity we share with the STAC corpus~\cite{asher:hal-02124399} appears to us as a great opportunity. Indeed, modeling natural language at discourse level is a task for which annotated resources are scarce. The Parallel Meaning Bank\footnote{\url{https://pmb.let.rug.nl/}} constitutes a notable exception and is annotated in DRT~\cite{kamp1981}. The STAC corpus is annotated in SDRT~\cite{asher2003logics}, which means that the proximity our two corpora share is an opportunity to study fine-grained modeling phenomena, as much regarding the correspondence between the two languages as for the logical models.

\paragraph{Ethics}

Recording during a game of \textit{Catan} allows us to capture long spontaneous interactions that (almost) do not contain personal data. \cite{amblard:hal-01079308,grouin2015possible} show that long interactions contain enough information to reduce the identification process to a (very) small amount of people. As the production of transcriptions of oral data is a very costly process, we want our corpus to be as widely sharable as possible. Therefore, following the recommendations of \cite{leidner2017ethical}, we decided to take care of ethical aspects by thinking them through the entire process of data collection and publication. The process was developed under the supervision and the validation of the Operational Committee for the Evaluation of Legal and Ethical Risks{\interfootnotelinepenalty10000 \footnote{\url{https://www.inria.fr/en/operational-committee-assesment-legal-and-ethical-risks}}} (OCELER) of the INRIA, in due respect of the General Data Protection Regulation\footnote{\url{https://gdpr-info.eu/}} (GDPR). 
All the participants signed an informed consent sheet, acknowledging they were giving us the right to record personal data (their voices) and share transcriptions of it. It was important to us to stress the fact that their consent was retractable at any point in the process. 
For now, we remove all mention of the names of the participants in the transcriptions and we publish transcriptions only. 

\section{General Presentation of DinG}

Dialogues in Games (DinG) is a corpus of manual transcriptions of real-life, oral, spontaneous multi-party dialogues between French-speaking players of \textit{Catan}\footnote{Copyright \textcopyright 2017 CATAN Studio, Inc. and CATAN GmbH. All rights reserved.}. \textit{Catan}, or \textit{Settlers of Catan}, is a board game for three to four players in which the main goal for each participant is to make their settlement prosper and grow, using resources that are scarce. Bargaining over these resources is a major part of the gameplay and constitutes the core of DinG's data.

\begin{example}
\label{ex:DinG-intro}
\textbf{A dialogue from DinG}
\item[Yellow$_1$] Tu veux bien me donner un mouton?
\vspace{-3mm}
\item[Yellow$_1$] Would you like to give me a sheep?
\item[Blue$_2$] Je veux bien euh un blé
\vspace{-3mm}
\item[Blue$_2$] I would like uh a wheat
\vspace{-1mm}
\item[Yellow$_3$] Et tu me donnes un mouton?
\vspace{-3mm}
\item[Yellow$_3$] And you give me a sheep?
\item[Blue$_4$] Et je te donne un mouton
\vspace{-3mm}
\item[Blue$_4$] And I give you a sheep
\end{example}

\textbf{Yellow} and \textbf{Blue} are players of \textit{Catan}, designated by the colour of their game tokens. In example~\ref{ex:DinG-intro}, \textbf{Yellow} and \textbf{Blue} are bargaining over resources: \textbf{Yellow} seeks a sheep; \textbf{Blue} offers to give them one, but only in exchange for wheat. \textbf{Yellow} secures the bargain in speech turn \textbf{Yellow}$_3$; \textbf{Blue} confirms. 
As the players have to speak to play, they do not discuss personal subjects outside the game setting, which makes it possible to completely anonymize the corpus by removing the players' names.

\paragraph{A game of \textit{Catan}} \index{Catan}

\begin{figure}[ht!]
    \centering
    \includegraphics[width=0.45\textwidth]{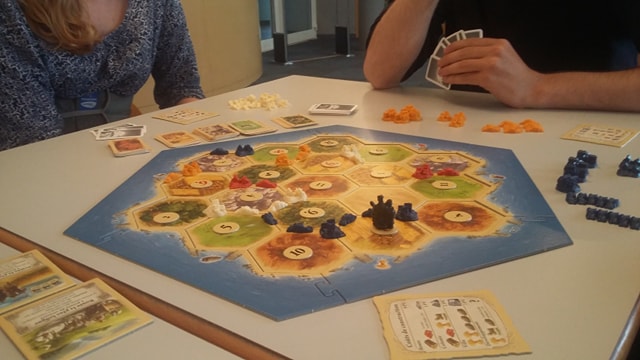}
    \caption[A photo of the game board during a game of \textit{Catan}]{The game board during a game of \textit{Catan}.}
    \label{fig:photo-catane}
\end{figure}

A typical game of \textit{Catan} takes place between 3 to 4 players and lasts between 30 and 90 minutes. The board is built using 19 hexagon terrain tiles of different colors: bright green pastures, yellow fields, grey mountains, brown carriers, dark green forests, and a desert tile. These tiles constitute the island of Catan, surrounded by water. See figure~\ref{fig:photo-catane} for an example. 

Most of the players we recorded for DinG never played \textit{Catan} before. Therefore, each recording was preceded by a phase of rules explanation conducted by an observer. After this phase, the rule book was handed to the participants for them to be able to play autonomously. 

\paragraph{Corpus Description}
\label{sec:DinG-description}

The corpus is composed of 10 recordings of games that last 70 minutes on average. The shortest recording is almost 40 minutes long (DinG8), the longest lasts a little over 1h44m (DinG1). Most of the recordings (all but n\textdegree 4, 5, and 6) were split into two parts because of a food break occurring during the game, as the recordings took place during university game nights. This division was kept in the transcription process, as it was easier for transcribers to work with shorter pieces of audio, as well as in the published data. The data presented in table~\ref{tab:DinG-data-plan-large} is computed on the merged recordings, one per game. 
We compute $CV$, the ratio of the standard deviation $\sigma$ to the mean $\mu$. A $CV < 100\%$ corresponds to a dataset with low variance.

\begin{table*}[t]
    \centering
    \begin{tabular}{c|r|r|r|r|r|r}
    \textbf{Name} & \textbf{Length}  & \textbf{Length} & \textbf{\# questions} & \textbf{\# turns} & \textbf{\# questions} & \textbf{\% questions}   \\
         & \textbf{(min)} & \textbf{(turns)} & &  \textbf{/minute} & \textbf{/minute} & \textbf{among turns} \\
         \hline 
         \hline
    DinG1 & \textbf{104.33} & \textbf{3,572} & \textbf{506} & 34.24 & 4.85 & \textbf{14.17} \\
    DinG2 & 86.31 & 2,969 & 290 & 34.40 & 3.36 & 9.77 \\
    DinG3 & 53.7 & 1,716 & 126 & 31.96 & 2.35 & \textbf{7.34} \\
    DinG4 & 75.93 & 2,985 & 333 & 39.31 & 4.39 & 11.16 \\
    DinG5 & 78.41 & 3,012& 362 & 38.41 & 4.62 & 12.02 \\
    DinG6 & 84.02 & 3,130 & 265 & 37.25 & 3.15 & 8.47 \\
    DinG7 & 96.34 & 3,293 & 340 & 34.18 & 3.53 & 10.32 \\
    DinG8 & \textbf{39.92} & 1,627 & 196 & \textbf{40.76} & \textbf{4.91} & 12.05 \\
    DinG9 & 41.71 & \textbf{795} & \textbf{69} & \textbf{19.06} & \textbf{1.65} & 8.68 \\
    DinG10 & 41.13 & \textbf{476} & \textbf{41} & \textbf{11.57} & \textbf{1.00} & 8.61 \\
    \hline 
    Global data & 701.8 & 23,575 & 2,528 & 33.59 & 3.60 & 10.72 \\
    CV & 34\% & 47\% & 57\% & 29\% & 40\% & 20\%
    \end{tabular}
    \vspace{1ex}
    
    \caption[DinG data -- observations per game, average on whole corpus and coefficients of variation]{DinG data -- observations per game, average on whole corpus and coefficients of variation ($CV$).}
    \label{tab:DinG-data-plan-large}
\end{table*}

DinG1 is the longest both with respect to time and amount of speech turns; it also contains the biggest amount of questions. While DinG9 and 10 are not the shortest in terms of time, their amount of speech turns and questions are significantly (more than 10\%) smaller than DinG8's (shortest in terms of time). This observation is supported by the fact that DinG9 and 10 present the smallest amount of speech turns per minute, while DinG8 presents the greatest: DinG8 lasts less time but DinG8's players talked at least twice more than DinG9 and DinG10's ones. Similarly, DinG8 presents the highest amount of questions per minute while DinG9 and DinG10 show the smallest ones.

The focus returns on DinG1 when we look at the percentage of questions among all the speech turns, as this game presents the highest percentage. The smallest percentage is shown by DinG3. DinG is homogeneous in terms of all the measures we considered in table~\ref{tab:DinG-data-plan-large}, as all the coefficients of variation stay under 60\%. While the amount of questions (identified as the utterances marked with a `?') varies quite a lot from one recording to another, the percentage of questions among turns stays very similar (around 10\%).

\section{Building the Corpus}
\label{sec:DinG-collection-process}

The recording part of the corpus collection took place during university game nights. 33 people participated in the recording process, 12 women and 21 men. All participants but 3 had a master's degree or higher. Each participant only appears once in the corpus. We collected as little personal data as possible, but we can say that the average age of the participants is around 25 years old, and all the participants are native French speakers. 

As we wanted the participants to feel as relaxed and natural as possible, the recordings were conducted in the room where the rest of the game night took place. Recording during the game nights raised some technical challenges, in particular, because different people were playing different games in the same physical space. Yet, it allowed us to record in a way that made the participants very comfortable: most of them report afterwards that they forgot the recording devices after the first fifteen minutes of playing. All recordings were conducted by a non-player observer, whose duties were to explain the experiment, find volunteers and supervise the smooth running of the process. 

\subsection{Data Processing}

Once a game is recorded, the raw audio file is given to transcribers. As our corpus was recorded in a noisy environment, transcribers had to pre-process the audio signal before starting to properly work on it. The preprocessing is done using Audacity\footnote{\url{https://www.audacityteam.org/}} and aims to reduce the peaks in the sound signal (corresponding to loud noises such as rolling the dices) in order to then be able to uniformly amplify the whole signal and make the voices clearer. These treatments diminish the background noise while sharpening the voices of the participants. The next step is manual transcription. 

Before choosing a transcription tool, we conducted a comparative study based on ergonomy, quality, and general characteristics of different specialized software. We also took into consideration the free availability of the software and their codes. We tested \textsc{ExpressScribe}\footnote{\url{https://www.nch.com.au/scribe}}, \textsc{Astali}\footnote{\url{http://ortolang108.inist.fr/astali/}}, \textsc{Youtube}\footnote{\url{https://support.google.com/youtube/answer/2734796?hl=en}}, \textsc{Transcriberjs}\footnote{\url{https://ct3.ortolang.fr/trjs/}}, \textsc{oTranscribe}\footnote{\url{https://otranscribe.com/}} and \textsc{elan}\footnote{\url{https://archive.mpi.nl/tla/elan}}. These software were evaluated on each of the following aspects: sound manipulation, navigation inside the recording, supported formats, text manipulation, transcription tools, speaker annotation, dialogue act annotation, management of overlaps, noises, and inaudible fragments. The tool that got the best evaluation was \textsc{elan}  \cite{wittenburg2006elan} because of its overall ergonomic design (for segmentation and transcription) in one tool while giving access to a visualization of the sound signal. Thanks to \textsc{elan}, DinG contains time-code alignment and disambiguation of speakers' overlaps.
Among the different possibilities that \textsc{elan} offers, we used mainly segmentation and transcription to produce the final version of DinG. 

\subsection{Process}
6 transcribers took part in the project, 5 of them were recruited among NLP students, one is an expert in production and synchronization of subtitles. They were trained for the task on a 5 minutes excerpt from DinG, that they all annotated and got to discuss with us and, when possible, between them. The transcribers were paid for the task. We counted 30 hours of work for the transcription of a recording of 1.5 hours. The transcribers followed different strategies, from minute-by-minute segmentation \& transcription in parallel to the full segmentation and then full transcription, speaker by speaker, of the whole recording.

\paragraph{Manual Segmentation in Speech Turns}
The first step in the process of getting the transcription is a manual segmentation of the recording in speech turns, called \textit{segments} in \textsc{elan}. We define a segment as a speech turn that is composed of a pseudo-sentence, an onomatopoeia, a noise, or any combination of the above. A speech turn is a theoretical linguistic unit corresponding to the verbal production of a speaker,~\cite{schegloff1974simplest}.
\textsc{Elan} allows us to process overlaps in a very simple and visual way. Thus, each segment constitutes a coherent linguistic unit.

\paragraph{Transcription} The transcription guide sets the norms to follow.
The guide is inspired by \cite{blanche1987franccais}, which has inspired many others transcription guides such as the one used within Transcriber\cite{Transcriber-LREC1998}. The \cite{blanche1987franccais} transcription guide advises not to use punctuation marks, so we explicitly added pauses duration and interrogative marks for our purposes, which makes us closer to guides such as Transcriber's one for French\footnote{\url{http://trans.sourceforge.net/en/transguidFR.php\#r241 }}. The main modifications are adaptations to the subject of our observation and the object of our research: (1) we specified the noise tags in order to adapt them to the board game context by adding tags such as [dice], [tokens]; (2) we added an explicit transcription of interrogative marks in order to account for utterances that were perceived (by the transcribers) as questions (rising intonation, answers given in the following dialogue turns). The transcription guide will be made available online. Furthermore, we produced a transcription and a segmentation, to preserve the dynamic aspect of interaction, for example by explicitly visualizing overlaps. Several automatic annotations were produced using SLODiM\footnote{\url{https://academia.slodim.fr/}}, in particular for disfluencies and syntax. 

As mentioned before, the transcribers who participated in the project have all received training on the same 5 minutes excerpt.
Everyone did an individual segmentation and transcription before pooling and comparing the results.
The inter-annotator agreement for transcriptions is calculated on the transcription of a 5 minutes excerpt of DinG2, pre-segmented, by two independent annotators (not working on the project before). They have received empty segments for the excerpt and filled them with transcriptions, following the transcription guide. 
First, we computed the agreements for the full transcriptions, see the first two lines of table~\ref{tab:interT-super-annot}. 
$\kappa_{ipf}$ is a modified version of Cohen's $\kappa$, computed using an \textit{iterative proportional fitting} algorithm, it includes the unmatched annotations in the agreement calculation. The raw agreement is computed by dividing the number of agreeing cases by the total number of cases~\cite{holle2015easydiag}.
It is important to stress that inter-annotator agreement on transcriptions is always low, as the amount of possible transcriptions is very large; yet, even taking this into account, the results we got were very low (under $0.3$). Then, we computed the agreements for the transcriptions from which we removed the noises and the pauses. This produced lines 3 and 4 of table~\ref{tab:interT-super-annot}, with results higher than $0.5$, which is usually considered to be a good agreement for transcriptions. This difference leads us to the conclusion that the quality of the recordings might be insufficient to grant an objective transcription of noises, on one hand, and also that transcriptions of the duration of pauses can vary from one transcriber to another.

\begin{table*}[t]
\centering
\begin{tabular}{|l|l|r|r|}
\hline
noise & unlinked/unmatched annotations& $\bm{\kappa_{ipf}}$ &  \textbf{Raw agreement} \\ \hline \hline
+  & +   & 0.28         & 0.28        \\ \hline
+ & -   & 0.28         & 0.28        \\ \hline \hline
- & +  & 0.52         & 0.55        \\ \hline
- & - & 0.53         & 0.55        \\ \hline
\hline
\multicolumn{4}{|l|}{After super-annotation}\\ 
\hline
\hline
+ & +  & 0.35         & 0.35        \\ \hline
+ & - & 0.35         & 0.35        \\ \hline
\end{tabular}

\caption[Interrater agreement for transcription before/after noise tags and pauses removal]
{Interrater agreement for transcription before/after noise tags and pauses removal, /after super-annotation, calculated with ELAN, following~\cite{holle2013modified}.}
\label{tab:interT-super-annot}
\end{table*}

\paragraph{Anonymization} It was of major importance for us to be able to distribute our resource while preserving the participants' private data. The last step in the transcription process is anonymization. Each of the players is identified with the colour of their game pieces: Red (\textbf{R}), White (\textbf{W}), Yellow (\textbf{Y}) or Blue (\textbf{B}). If a name is pronounced out loud, it is replaced in the transcription by the name of the corresponding color, in upper case (e.g. ``your turn, BLUE''). Outside noises and speakers are assigned to an outside speaker called Other (\textbf{O}).

\paragraph{Super-Annotatation} Once the transcription is done, it is given to a super-annotator, whose goal is to proofread, homogenize the corpus via the correction of typos, standardization of the noise tags, of the transcription of onomatopoeias, of the writing of numbers\footnote{In French, « un » corresponds both to ``a'' and to ``one''. The guide asks for disambiguation, as much as possible. Following this idea, we transcribe « 1 », « 2 », etc. for amounts and « un » when the usage corresponds to the determiner.} and of the anonymization. Table~\ref{tab:interT-super-annot} shows the raise in transcriptions quality obtained after super annotation, computed on non-specialists' transcriptions: the last two lines correspond to the results obtained by calculating the agreement on the proofread transcriptions.

\paragraph{The Corpus}
The corpus is available on Gitlab\footnote{
\url{https://gitlab.inria.fr/semagramme-public-projects/resources/ding/}
}.
It is distributed under the Attribution ShareAlike Creative Commons license (CC BY-SA 4.0). Each game is available as a numbered .txt file, exported from \textsc{elan}. We export the transcriptions as \texttt{Traditional Transcript Text} in order to generate text files that would be both readable by human observers and could easily be manipulated by scripts.

The files correspond to linearised versions of the games. Each segment appears on one line, that starts by the number of this segment in the transcription, then the letter that identifies the speaker (\textbf{R}, \textbf{W}, \textbf{Y}, \textbf{B} or \textbf{O}), then the transcription of what has been said. The next line contains the time codes of the beginning and end of the segment. When two following speech turns do not overlap, the gap between the two is calculated automatically and written in brackets on the next line. These elements are shown in Figure~\ref{ex:verbatim-DinG}, see in particular the gap between \texttt{009} and~\texttt{010}.

\begin{figure} \ \

\normalfont{
\begin{verbatim}
009 Y   j'aimerais bien faire 7 
        pour une fois
    00:00:14.438 - 00:00:15.880
    (0.64)

010 R   en fait t'as (te-) t'étai
        s contente parce que juste 
        tu as fait un double 6 et 
        qu'en général c'est cool 
        dans les jeux [rire]
    00:00:16.518 - 00:00:21.910

011 Y   ouais c'est ça
    00:00:21.712 - 00:00:22.718

012 R   [rire]
    00:00:21.915 - 00:00:23.219
\end{verbatim}
}
\normalfont{
\begin{description}
\item[009 Y] I would like to get a 7 for once
\item[010 R] in fact your have (y-) you were happy because simply you got a double 6 and generally it's cool in games [laugh]
\item[011 Y] yeah that's it
\item[012 R] [laugh]
\end{description}
}
\caption{Excerpt from DinG transcription and translation, DinG6.}
\label{ex:verbatim-DinG}
\end{figure}

\section{Question Annotation}

As we have indicated, the objective of the development of this resource is to make available long dialogues in French, but also to study their functioning. As dialogues are characterized by the use of questions and answers, we focused on analyzing them in DinG. As the questions were explicitly transcribed through question mark, we were able to automatically retrieve all of them along with a small context with the two preceding and the two following utterances. The questions were first automatically assigned tags, then annotated by hand to correct and augment this first annotation. 

\subsection{Annotation Scheme}
Several annotation schema for different dialogue phenomena have been developed over the years.
Following the developments we presented in the previous sections, we annotated the questions from DinG using an annotation schema for questions adapted from~\cite{blandon2019toward}, presented in table~\ref{tab:LAWXIII-Q-tags}.

Our annotation schema is an easy-to-use one. We would like to extend the annotation of questions and answers through more detailed annotation schemas such as the ones developed for STAC~\cite{asher:hal-02124399} or inspired by insights from~\cite{bazillon-etal-2011-qui,abeille2013french,smirnova2021question}. 

Our beginning assumption is that the corpora would contain at least two well-known and well-defined categories of questions: \textit{yes/no}-questions and \textit{wh}-questions. Some questions are similar to \textit{wh}-questions or \textit{yes/no}-questions in usage but have a different form: e.g., \textit{wh-in-situ} questions such as ``You saw what?'', or \textit{yes/no}-questions without inversion such as ``You saw him?''. We decided not to introduce new categories for these based on their semantics and pragmatics.

Some questions containing a disjunction (e.g. ``Do you go on Monday or on Tuesday?'') are semantically and pragmatically similar to \textit{wh}-questions, but are syntactically closer to \textit{yes/no}-questions. This kind of question exhibits subject-auxiliary inversion (in English) but does not ask for the confirmation or denial of the proposition that it expresses. Instead, it expects the addressee to provide some missing information within the set of options to choose from. We call this type of questions \textit{disjunctive questions}.

\begin{table}[ht!]
    \centering
    \begin{tabular}{c|l}
    \textbf{Tag} & \textbf{Name} \\
    \hline
    YN & yes/no-question \\
    WH & \textit{wh-}question \\
    DQ & disjunctive question \\
    CS & completion suggestion \\
    PQ & phatic question \\
    N/A & non-assigned
    \end{tabular}
    \caption[Set of question tags]{Question tags.}
    \label{tab:LAWXIII-Q-tags}
\end{table}

Some questions have the syntactic characteristics of a \textit{yes/no}-question or a \textit{wh}-question, but are used with different pragmatics and/or semantics. For example, the speaker of the question can suggest a way to complete the utterance of the previous speaker, and the expected answer would confirm or deny this suggestion. This is subtly different from a prototypical \textit{yes/no}-question because the speaker of the question does not necessarily ask their interlocutor to confirm the truth value of the semantic content of the suggestion. We call these types of questions \textit{completion suggestions}.

Other questions take the appearance of a \textit{yes/no}-question or a \textit{wh}-question, respectively, but the context and intonation of the utterance make clear that the speaker is not actually interested in the confirmation or denial of the proposition.  Instead, such questions can have various so-called \textit{phatic} functions, i.e. their semantic content is less important than their social and rhetorical functions \cite{FREED1994621,senft2009}). We call this type of questions \textit{phatic questions}.

After first experiments with human annotators, we added a last category of automatic annotation: questions that we cannot assign a category to: \texttt{N/A}. It is used in particular if the complete utterance is « (xxx) ? », which is interpreted as ``the person who was transcribing could not figure out any words but still picked up an interrogative/rising intonation''.

\subsection{Automatic Annotation}
\label{sec:auto-annot-DinG}
As phatic questions and completion suggestion are categories that are highly context-dependant, we did not pre-annotate them automatically. 
Some of the utterances contain multiple interrogative marks, corresponding to several questions asked in a row, without pauses between them, with several rising intonation points. The automatic annotation only annotated once.

The automated annotation of questions from DinG, called hereafter \textit{utterance number $n$}, follows the next rules, in this order:
\begin{enumerate}
    \item \label{itm:YN} If the utterance number $n+1$ is affirmative (starts with the word « oui », « ouais » or « ok ») or negative (starts with the word « non »), the question in utterance $n$ is automatically tagged as a \textit{yes/no}-question: \texttt{YN}.
    \item \label{itm:WH} If the utterance contains a French \textit{wh}-word
    , the question is automatically tagged as a \textit{wh-}question: \texttt{WH}. \cite{boritchev2021picturing} gives a list of \textit{wh-}words in French. 
    \item \label{itm:DQ} If the utterance contains « ou », the question is automatically tagged as a disjunction: \texttt{DQ}. 
    \item \label{itm:N/A} \texttt{N/A} otherwise.
\end{enumerate}

The automatic annotation was able to assign a tag to $772$ out of $2504$\footnote{Note that $24$ questions were removed from the process for technical issues.} utterances containing at least one interrogation point, which corresponds to $\sim 31 \%$. 

The automatic annotation was systematically wrong in several cases:
    i) Several times, a \textit{wh-}question is directly followed by a negation (see example~\ref{ex:annot-WHwNo}). Following \ref{itm:YN}. in the automatic annotation rules, the question was assigned the \texttt{YN} tag in this case, while it should have been a \texttt{WH}.
    ii) A lonely « quoi ? » (``what?'') is most of the time phatic (if it is the only content of the utterance). Following \ref{itm:WH}. in the automatic annotation rules, the question was assigned the \texttt{WH} tag in this case, while it should have been a \texttt{PQ}.

\begin{example}

\textbf{\textit{Wh-}question followed by a `no', DinG4 }\\
\label{ex:annot-WHwNo}

\ExLi{W}{1150}{\textbf{qui veut une pierre contre un mouton?}}
{\textbf{who wants a rock for a sheep? }}
\ExLi{R}{1151}{\textbf{non} mais vraiment}{\textbf{no} but really}
\ExLi{R}{1152}{oui oui oui je euh je prends totalement}{yes yes yes yes I uh I'll totally take [it]}
\end{example}

A way to interpret example (\ref{ex:annot-WHwNo}) is that \textbf{R} started the utterance \textbf{R}$_{1151}$ without listening to \textbf{W}$_{1150}$, to answer \textbf{B}$_{1149}$. This hypothesis is supported by the content of \textbf{R}$_{1152}$, which is an answer to \textbf{W}$_{1150}$. 
This has to do with multi-thread conversation phenomena that we do not take into account in this annotation.

\subsection{Human Annotation}

The annotations was conducted by $10$ people, among which $3$ did the full annotation and $7$ annotated subparts of the corpus. All the annotators but $2$ are native speakers of French. Annotating the whole corpus took 6 hours to one human annotator, which is why most of the annotators only went through part of the task. 

The annotations were performed using spreadsheets. Part of the questions were automatically pre-annotated, through the process presented in section~\ref{sec:auto-annot-DinG}. 
\begin{figure}[ht!]
    \centering
    \includegraphics[width=0.45\textwidth]{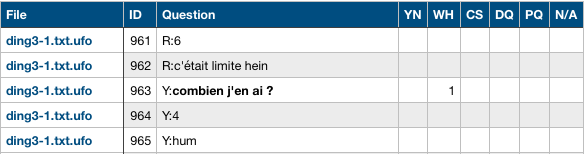}

\textbf{R:} 6
\textbf{R:} it was borderline eh
\textbf{Y:} \textbf{how many do I have?}
\textbf{Y:} 4
\textbf{Y:} hum

(a)

    \includegraphics[width=0.45\textwidth]{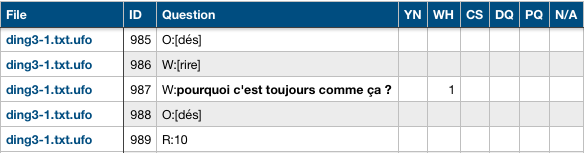}

\label{ex:EN-ex-annot-auto-DinG-PQ}
\textbf{O:} [dice]
\textbf{W:} [laugh]
\textbf{W:} \textbf{why is it always like that?}
\textbf{O:} [dice]
\textbf{R:} 10

(b)
    \caption[Automatic annotation of DinG -- \textit{wh}-question]{Screen-shots and translations of the spreadsheet used to annotate questions from DinG with an automatic annotation of the central utterance.}
    \label{fig:ex-annot-auto-DinG-WH}
    \label{ex:EN-ex-annot-auto-DinG-WH}
\end{figure}

In example~(a), figure~\ref{fig:ex-annot-auto-DinG-WH}, \textbf{Yellow} is asking a \textit{wh-}question introduced by the French \textit{wh-}word « combien » (``how much''). The \textit{wh-}word is identified by the automatic tagger, the `1' is put in the \texttt{WH} column. The human annotator can validate this tagging by not changing it and proceeding to the next question. 

In example~(b)
, figure~\ref{fig:ex-annot-auto-DinG-WH}, \textbf{White} is asking a \textit{wh-}question introduced by the French \textit{wh-}word « pourquoi~» (``why''). The \textit{wh-}word is identified by the automatic tagger, the `1' is put in the \texttt{WH} column. Yet, the human annotator can see, from the surrounding context, that \textbf{White}'s question, while \textit{wh-} in its form, is actually phatic. \textbf{White} is in fact (fake?) complaining about something related to the game, most likely the result of the roll of the die. The human annotator needs to correct this tagging by moving the `1' to the \texttt{PQ} column.

\subsection{Results of the Human Annotation}
\paragraph{Consequences of the Annotation}
The human annotators encountered several difficulties throughout the annotation. Annotator~1 was the first to conduct the full annotation. Their experience led us to implement the following decisions in the next annotations.
\begin{itemize}
    \item Annotate question-tags (« j'ai fini, c'est ça ? » / ``I finished, isn't it?'') as \texttt{YN}.
    \item 
    Separate the « ou pas » (or not) question tag from the others and tag questions finishing by « ou pas » as disjunctive, because they contain an ``or'' (« ou »). This decision is arguable and could be replaced by a tagging of this type of questions as \texttt{YN}.
    \item We annotate as phatic, unless clear from context that it's not, the following questions:  short questions such as « quoi ? » (``what?''), « c'est bon ? » (``all good?''), rhetorical/theatrical questions such as « sérieusement ? » (``seriously?''), « encore ? » (``again?''), « pourquoi c'est encore le 7 ? » (``why is it again the 7?'').
    \item A question in speech turn $n$ is tagged as a completion suggestion only if the speech turn $n-1$ is uttered by a speaker different from $n$'s one. This comes from the fact that if a person pauses in the middle of a question, it can be split into two speech turns and create the false impression of a completion suggestion.
\end{itemize}

These decisions eased the work of the following annotators, however, they didn't solve all the difficulties. The following section presents some of the cases that
raised difficulties during the annotation process.

\paragraph{Annotation Results}
The annotation results obtained by the three annotators who worked on the whole corpus are presented in table~\ref{tab:results-anno-DinG}. The inter-annotator agreement scores are shown in table~\ref{tab:kappa-DinG}. Annotator~1 was the first to annotate the whole corpus; after they turned in their annotation, the annotation guide was adjusted, as presented above. In particular, the definition of completion suggestions (CS) in the context of DinG was clarified. This explains the fact that Annotator 1 tagged no questions as CS, while the other found a few. 

\begin{table*}[ht!]
\centering
\begin{tabular}
{l|rrrrrrr}
                                & \multicolumn{1}{c}{\textbf{YN}} & \multicolumn{1}{c}{\textbf{WH}} & \multicolumn{1}{c}{\textbf{CS}} & \multicolumn{1}{c}{\textbf{DQ}} & \multicolumn{1}{c}{\textbf{PQ}} & \multicolumn{1}{c}{\textbf{N/A}} & \multicolumn{1}{c}{\textbf{Total}} \\
                                \hline
\textbf{Automatic}              & 494                             & 262                             & -                               & 16                              & -                               & -                                & 772                                \\
\hline
\textbf{Annotator 1
}   & 1,588                           & 608                             & 0                               & 71                              & 89                              & 100                              & 2,456                              \\
\textbf{Annotator 2
}    & 1,389                           & 602                             & 17                              & 115                             & 364                             & 26                               & 2,513                              \\
\textbf{Annotator 3
} & 1,345                           & 572                             & 7                               & 106                             & 458                             & 23                               & 2,511                              \\
\hline
\textbf{Average amount}             & 1,441                           & 594                             & 8                               & 97                              & 304                             & 50                               & 2,493                              \\
\textbf{Average percentage}     & 57.78 \%                        & 23.82\%                         & 0.32 \%                         & 3.90 \%                         & 12.18\%                         & 1.99\%                           & 100\%                             
\end{tabular}

\caption[Annotation results for DinG's questions annotations]{Annotation results for DinG's questions annotations of the 3 annotators who did all annotations.}
\label{tab:results-anno-DinG}
\end{table*}

It is also interesting to note that the annotators did not annotate the same amount of questions even though they were all working with the same corpus. Our hypothesis is that this results from the utterances that contain two interrogation points: even though the convention was to annotate each of the questions separately, some examples seemed to be open to interpretation. Example~\ref{ex:annot-two-?} shows an utterance transcribed with two interrogation marks. \textbf{R}$_{1074}$ was attributed two tags: a `1' in the YN column, because the expected answer for the ``and some clay?'' part of the utterance is a `yes' or a `no'; and a `1' in the PQ column, as the second part of the utterance is used for metacommunication purpose, to put an emphasis on the first part or perhaps on the fact that the consent of the addressee matters. 

\begin{example}\textbf{Utterance transcribed with two `?', DinG2}

\label{ex:annot-two-?}
\ExLi{Y}{1073}{1 mouton et autre chose oui}{1 sheep and something else yes}
\ExLi{R}{1074}{\textbf{et de l'argile ? ça te va ?}}{\textbf{and some clay? is that okay for you?}}
\ExLi{Y}{1075}{ça me va parfaitement}{it is perfectly okay for me}
\end{example}

In general, it seems that tagging the questions as polar or \textit{wh-} is an easier task than assigning the \texttt{CS}, \texttt{DQ}, or \texttt{PQ} tags. \texttt{CS} and \texttt{PQ} are categories that correspond to the pragmatics of dialogue, they are highly open to interpretation. The case of disjunctive questions is more interesting: they constitute on average less than $4\%$ of the corpus, and they are quite likely to be confused with \textit{yes/no}-questions because of a frequently encountered mixed form such as the one in example~(\ref{ex:annot-YNwDQ}).

\begin{example}
\textbf{\textit{Yes/no}-question with an embedded disjunction, DinG1}\\

\label{ex:annot-YNwDQ}
\ExLi{R}{569}{non moi je je}{no I I I}
\vspace{-1mm}
\ExLi{R}{570}{\textbf{j'achète du mouton quelqu'un veut du (1s) blé ou du bois?}}{\textbf{I'm buying sheep does anyone want (1s) wheat or wood?}}
\vspace{-1mm}
\ExLi{B}{571}{non}{no}
\end{example}

In example~(\ref{ex:annot-YNwDQ}), \textbf{R}$_{570}$ starts as a \textit{yes/no}-question with a do-support (``does anyone want''), but continues with a disjunctive part (``wheat or wood?''). The decision was taken to follow the top-most form and thus annotate this type of question with the \texttt{YN} tag.

\paragraph{Inter-annotator Agreement}
Table~\ref{tab:kappa-DinG} presents the inter-annotator agreement scores for the three annotators that annotated all of the questions. As Cohen's $\kappa$ measures agreement between two annotators only, we also computed Fleiss' $\kappa$ for all three annotators. All the scores are quite high ($\kappa > 0.61$), but it is particularly interesting to notice that the agreement between annotator 2 and annotator 3 is higher than $0.8$. Annotators 2 and 3 performed the annotation after the aforementioned modifications of the annotation guidelines, inspired by annotator 1's experience.

\begin{table}[ht!]
\centering
\begin{tabular}{c|rr}
             & Average Cohen $\kappa$ & Fleiss $\kappa$ \\
             \hline
A1 + A2      & 0.651                  & -           \\
A1 + A3      & 0.615                  & -           \\
A2 + A3      & \textbf{0.804}         & -           \\
A1 + A2 + A3 & -                  & 0.693          
\end{tabular}
\caption[Inter-annotator agreement scores for DinG's questions annotations]{Inter-annotator agreement scores for DinG's questions annotations, where A1, A2 and A3 are the three annotators that annotated all of the questions.}
\label{tab:kappa-DinG}
\end{table}

Three annotators annotated only the first half of the questions. Partial annotator~2 annotated in parallel with annotator~1, so the adjustments in the annotation guide mentioned above took place after partial annotator~2 turned their annotation in. Table~\ref{tab:kappa-DinG-half} presents the inter-annotator agreement scores for the three partial annotators. All the scores are high ($\kappa > 0.77$), but it is particularly interesting to notice that the agreement between partial-annotator 1 and partial-annotator 3 is higher than $0.87$. Partial-annotators 1 and 3 performed the annotation after the aforementioned modifications of the annotation guidelines, inspired by annotator 1's experience, while partial-annotator 2 annotated with the same guidelines as annotator 1. 

\begin{table}[ht!]
\centering
\begin{tabular}{c|rr}
             & Average Cohen $\kappa$ & Fleiss $\kappa$ \\
             \hline
P1 + P2      & 0.797                  &  -         \\
P1 + P3      & \textbf{0.872}                  &  -          \\
P2 + P3      & 0.771                  & - \\
P1 + P2 + P3 & -                   & 0.813          
\end{tabular}
\caption[Inter-annotator agreement scores for the first half of DinG's questions annotations]{Inter-annotator agreement scores of the annotator which annotated the first half of DinG's questions. }
\label{tab:kappa-DinG-half}
\end{table}

\section{Conclusion}

The next step for questions in DinG is to produce a golden version. 
It will straightforwardly contain all the annotations for which the 6 annotators agree (from the first half of the questions). Then, we need to examine the annotations from the first half of the questions for which all annotators but A1 and P2 agree. In practice, we are waiting for more complete annotations before building and publishing the gold corpus.

Another aspect that we could not develop in this article due to lack of space is the comparison with other existing resources for French on the one hand, and for the Catane game on the other hand. For the French resources, the comparison with a large corpus such as ESLO\footnote{\url{http://eslo.huma-num.fr/}}~\cite{eshkol2011grand} shows important similarities, with a smaller volume of questions. Another comparison is with the French QuestionTreebank\footnote{\url{http://alpage.inria.fr/Treebanks/FQB/}} (FQB,~\cite{seddah2016hard}), the reference corpus for questions in French. This corpus is built from governmental websites' FAQs. We expect to find multiple differences because DinG is intended to highlight spontaneous production. Finally, the comparison with the STAC corpus remains an important step for the analysis of the interaction dynamics.

In future work, we are considering anonymizing the oral data, following approaches such as \cite{qian2017voicemask}. When we manage to do so, we will contact again the participants as they will have to sign a new consent for their data to be published. 
This step is very time-consuming, but can be done in parallel with the linguistic analysis. However, getting this data will increase the interest of the resource and its dissemination.

\section*{Acknowledgements}
This work was supported partly by the french PIA project ``Lorraine Universite d’Excellence'', reference ANR-15-IDEX-04-LUE. We thank immensely Esteban Marquer, Julien Botzanowski, Laurine Jeannot, Léa Dieudonat, Srilakshmi Balard, Lucille Dumont, Amandine Lecompte and Samuel Buchel for their great transcription work.
We thank greatly Amandine Decker, Bruno Guillaume,  Pierre Lefebvre, Chuyuan Li, Léo Mangel, Siyana Pavlova, Guy Perrier, and Valentin Richard for their time and their annotation to our work.
We also thank our reviewers for all the useful comments and references.
\section{Bibliographical References}\label{reference}

\bibliographystyle{lrec2022-bib}
\bibliography{lrec2022-example}

\end{document}